\documentclass{article}

\usepackage{PRIMEarxiv}

\usepackage[utf8]{inputenc} 
\usepackage[T1]{fontenc}    
\usepackage{hyperref}       
\usepackage{url}            
\usepackage{booktabs}       
\usepackage{amsfonts}       
\usepackage{nicefrac}       
\usepackage{microtype}      
\usepackage{lipsum}
\usepackage{fancyhdr}       
\usepackage{graphicx}       
\usepackage{array}
\graphicspath{{media/}}     

\title{CDJUR-BR - A Golden Collection of Legal Document from Brazilian Justice with Fine-Grained Named Entities}

\author{
  Maurício Brito, Vládia Pinheiro, Vasco Furtado,\\
  João Araújo Monteiro Neto,\\
  Francisco das Chagas Jucá Bomfim e André Câmara Ferreira da Costa\\
  Affiliation\\
  Universidade de Fortaleza\\
  Fortaleza - BR\\
   \And
  Raquel Silveira\\
  Affiliation\\
  Instituto Federal de Educação, Ciência e Tecnologia do Ceará\\
  Fortaleza - BR\\
     \And
  Nilsiton Aragão\\
  Affiliation\\
  Tribunal de Justiça do Estado do Ceará\\
  Fortaleza - BR\\
}

\begin{document}
\maketitle

\begin{abstract}
A basic task for most Legal Artificial Intelligence (Legal AI) applications is Named Entity Recognition (NER). However, texts produced in the context of legal practice make references to entities that are not trivially recognized by the currently available NERs. There is a lack of categorization of legislation, jurisprudence, evidence, penalties, the roles of people in a legal process (judge, lawyer, victim, defendant, witness), types of locations (crime location, defendant's address), etc. In this sense, there is still a need for a robust golden collection, annotated with fine-grained entities of the legal domain, and which covers various documents of a legal process, such as petitions, inquiries, complaints, decisions and sentences. In this article, we describe the development of the Golden Collection of the Brazilian Judiciary (CDJUR-BR) contemplating a set of fine-grained named entities that have been annotated by experts in legal documents. The creation of CDJUR-BR followed its own methodology that aimed to attribute a character of comprehensiveness and robustness. Together with the CDJUR-BR repository
we provided a NER based on the BERT model and trained with the CDJUR-BR, whose results indicated the prevalence of the CDJUR-BR.
\end{abstract}


\section{Introdução}
Processamento de Linguagem Natural (PLN) permite  a manipulação automática, rápida e eficiente de grande volume de  documentos textuais. Em vários domínios, como o jurídico, o impacto do PLN é importante, pois permite automatizar completa ou parcialmente tarefas como classificação de processos \cite{peixoto2020projeto}, sumarização de documentos \cite{kanapala2019text,yamada2019building}, geração automática de sentenças e pareceres \cite{zhong2020does} e  buscas de jurisprudências e normas jurídicas \cite{angelidis2018named,boella2019semi}. Uma tarefa básica para a maioria das aplicações em Inteligência Artificial Jurídica (Legal Artificial Intelligence - Legal AI, em inglês) é a de Reconhecimento  de  Entidades  Nomeadas  (REN). Para além de mera classificação gramatical, a tarefa de REN busca identificar e qualificar se um trecho do texto se refere a entidades como pessoas, locais, organizações, datas, dentre outras, agregando ao texto informação semântica \cite{yadav2019survey}. A literatura cientifíca e as bibliotecas de código de IA estão repletas de reconhecedores de entidades genéricas, como as mencionadas anteriormente \cite{schmitt2019replicable,li2020survey}, que são treinados a partir de coleções de textos rotulados por humanos especialistas e coletados de repositórios diversos como enciclopédias, jornais de notícias ou da literatura ficcional ou não-ficcional. Esta coleção de referência contendo textos anotados com informação extraordinária denomina-se comumente de Coleção Dourada ou, em inglês, \textit{Golden Collection} \cite{schmitt2019replicable,jiang2016evaluating,atdaug2013comparison}. 
 
 No entanto, textos de domínios específicos, como os produzidos no contexto da prática jurídica, fazem referências a outras entidades que não são trivialmente reconhecidas pelos RENs disponíveis atualmente. Isso se dá pelo fato de que textos jurídicos possuem discurso com termos técnicos, racionalmente ordenado e objetivando uma comunicação clara, precisa e concisa \cite{coan2003atributos}. É através do texto que o autor explicita a sua pretensão jurídica quando elabora uma demanda judicial, daí ser comum referências à entidades como leis e doutrinas, réus, vítimas, testemunhas, penalidades, etc, que embasam e buscam dar clareza às peças processuais. Da mesma forma, as respostas às petições legais, sentenças e decisões são produzidas seguindo similar vocabulário, estrutura e referências a estas entidades. Como em toda área de conhecimento, certas entidades requerem conhecimento técnico do anotador para serem devidamente rotuladas em uma coleção dourada. Na área jurídica não é diferente. A habilidade de rotular que uma determinada citação (e.g. \textit{artigo 5. da CF}) é uma norma legal, muito embora não seja de natureza complexa, necessita de conhecimento e experiência na produção dos documentos. Mais desafiador ainda se torna classificar a relevância da mesma para caracterizar do que se trata o texto jurídico. Trata-se de uma \textit{norma principal} e que define o assunto da petição? Ou se trata de uma \textit{norma acessória} servindo somente para apoiar os argumentos do peticionante? Outro exemplo seria o de classificar o papel semântico das pessoas mencionadas na peça processual: é uma vítima? o réu? o juiz? As respostas a essas perguntas são de natureza interpretativa e requer do anotador conhecimento técnico para fornecê-las.
 
 O contexto supramencionado contribui para que sejam raros os exemplos de extensas coleções douradas para o domínio jurídico \cite{de2018lener,leitner2020dataset}, o que é um obstáculo para o desenvolvimento de aplicações em \textit{Legal AI}.  Para a língua portuguesa,  os poucos exemplos existentes somente possuem entidades básicas rotuladas. Há carência de categorizações de legislação, jurisprudência, provas, penalidades, dos papéis das pessoas em um processo jurídico (juiz, advogado, vítima, réu, testemunha), dos tipos de locais (local do crime, endereço do réu), etc. Neste sentido, persiste ainda a necessidade de uma coleção dourada robusta, anotada com entidades refinadas do domínio jurídico, e que abranja diversos documentos de um processo legal, como petições, inquéritos, denúncias, decisões e sentenças. Neste artigo, descrevemos o desenvolvimento da Coleção Dourada do Judiciário Brasileiro (CDJUR-BR) contemplando um conjunto de entidades nomeadas anotadas de forma manual por especialistas em documentos jurídicos.  A criação da CDJUR-BR seguiu uma metodologia própria que visou atribuir o caráter de abrangência e robustez à coleção contendo 21 entidades refinadas (sob a perspectiva dos especialistas envolvidos) e que possa servir ao processo de treinamento e validação de modelos de \textit{Legal AI} para língua portuguesa. Especialmente, para a rotulação das normas legais e seus artigos foram aplicadas etapas adicionais de refinamento e validação, pois estas entidades, por se tratarem da formalização do raciocínio jurídico, são consideradas essenciais a uma série de aplicações do PLN no domínio jurídico.
Para o desenvolvimento da metodologia e a realização dos trabalhos de anotação de entidades, foi criada uma força tarefa de especialistas de diferentes perfis vindos de três instituições fundamentais à Justiça, como o Ministério Público e Tribunal de Justiça do Estado do Ceará, Brasil.

Este trabalho têm como contribuições apresentar a metodologia própria desenvolvida para guiar as atividades de anotação manual de Entidades Nomeadas Refinadas, e disponibilizar a CDJUR-BR, uma coleção padrão-ouro, produzida para treinamento e validação de algoritmos de aprendizado
de máquina utilizados em soluções de Legal AI. Para isso, foram realizados experimentos visando demonstrar que o conjunto de dados criado suporta o desenvolvimento de REN eficazes em documentos legais. Juntamente com o repositório da CDJUR-BR provemos adicionalmente um REN baseado no BERT e treinado a partir da CDJUR-BR. A acurácia do REN, em comparação com outros reconhecedores (quando essa comparação foi possível de ser realizada), mostrou as vantagens e prevalência do uso da CDJUR-BR.

\subsection{Problemática e Questões de Pesquisa}
\label{sec:problemática-questoes-pesquisa}

Durante a realização da revisão bibliográfica, foi constatado que existem poucos \textit{corpus} linguísticos computacionais disponíveis em língua portuguesa para dar suporte às aplicações de \textit{Legal AI}. As coleções existentes são limitadas e, em grande parte, feitas a partir de \textit{corpus} de notícias ou retirados de \textit{sites} de conhecimentos gerais como a wikipédia. São raros os conjuntos de dados voltados para o domínio jurídico. Além disso, foi observada a ausência de informações sobre as metodologias de anotação utilizadas. Diante dessa realidade, este artigo propõe a elaboração de uma metodologia própria para as anotações manuais de documentos que compõem as peças de um processo jurídico, e faz uso prático da mesma para criar um coleção dourada de entidades nomeadas para o judiciário brasileiro.

Diante desta problemática, as seguintes questões de pesquisa são levantadas para direcionar o desenvolvimento deste trabalho:

\begin{itemize}
    \item \textit{QP1} - Como elaborar uma metodologia de anotações manuais de entidades nomeadas que contemple as especificidades e complexidades do domínio jurídico?
    \item \textit{QP2} - A coleção dourada gerada é adequada para o treinamento e validação de modelos de \textit{Legal AI}?
\end{itemize}

O restante deste artigo está estruturado da seguinte forma: Na seção 2, apresentamos os trabalhos relacionados. Na seção 3, descrevemos a metodologia de anotação, as técnicas e as ferramentas usadas. Na seção 4, explicamos a avaliação  da CDJUR-BR na tarefa de Reconhecimento de Entidades Nomeadas e apresentamos os resultados dos experimentos. Por fim, as conclusões e recomendações para trabalhos futuros.

\section{Trabalhos Relacionados}
\label{sec:trabalhos-relacionados}

A prática de criar coleções douradas para o contexto jurídico tem na Europa, seus maiores exemplos. O estudo realizado por \cite{leitner2020dataset}, desenvolveu um conjunto de dados, em alemão, de entidades nomeadas e expressões temporais, a partir de 750 documentos de decisões judiciais publicadas online pelo Ministério Federal da Justiça e Defesa do Consumidor da Alemanha. Este conjunto de dados é parte de um esforço da União Europeia (UE) para apoiar, especialmente, PMEs que desejam atuar em outros mercados da UE, oferecendo serviços relacionados à compliance. No processo de criação do dataset, 54.000 entidades foram anotadas manualmente, mapeadas para 19 classes semânticas (pessoa, juiz, advogado, país, cidade, rua, paisagem, organização, empresa, instituição, tribunal, marca, lei, decreto, norma jurídica europeia, regulamento, contrato, decisão judicial e literatura jurídica). O artigo não apresenta metodologia para a etapa de anotação, mas descreve que foram desenvolvidas instruções específicas para as anotações. Estas instruções foram utilizadas para que um segundo anotador realizasse marcações em uma parte, não especificada, dos documentos. Para os documentos que tiveram duas anotações foi alcançada a concordância entre anotadores de 0,89 no coeficente Kappa. Eles também relatam que alcançaram a melhor pontuação F1 de 95,46 com um modelo  de rede neural BiLSTM. 

No trabalho desenvolvido por \cite{angelidis2018named}, foram anotadas 254 partes do Diário do Governo Grego, relativos a leis, decretos presidenciais, decisões ministeriais, regulamentos, como também os assuntos referentes a decisões relacionadas ao planejamento urbano, rural e ambiental entre os anos de 2.000 e 2.017. As anotações envolveram 6 tipos de entidades: PESSOA, para qualquer nome de pessoa citada nos documentos; ORGANIZAÇÃO, para qualquer referência a organização pública ou privada; ENTIDADE GEOPOLÍTICA, para qualquer referência a uma entidade geopolítica (por exemplo, país, cidade, unidade administrativa grega, etc.); MARCO GEOGRÁFICO, para especificar referências a entidades geográficas como bairros, estradas, praias, que constam principalmente de regulamentações relativas a planejamentos urbanísticos e topográficos; REFERÊNCIA À LEGISLAÇÃO, qualquer referência à decretos presidenciais, leis, decisões, regulamentos e diretivas da União Europeia ou Grega; REFERÊNCIA A DOCUMENTOS PÚBLICOS, qualquer referência a documentos ou decisões que tenham sido publicadas por uma instituição pública que não sejam consideradas uma fonte primária de legislação. O objetivo deste trabalho foi o reconhecimento de entidades nomeadas (REN) para enriquecer  um  grafo  de  conhecimento da legislação grega com informações mais detalhadas sobre as EN descritas. O artigo não menciona as atividades realizadas no processo de anotação manual das EN. Nos experimentos realizados, eles relatam que alcançaram a Média Macro para F1-Score de 0,88 como avaliação do Modelo REN desenvolvido.

Huang et al. \cite{huang2020named} realizaram o reconhecimento de entidades nomeadas para documentos de julgamento chineses com base nos modelos BiLSTM e CRF,  obtendo,  no  geral,  a  pontuação  de  75,35  de  F1.  Para  tanto,  precisaram trabalhar  com  as  particularidades  do  idioma  chinês  que  não  há  limites  óbvios entre as palavras (como nas línguas ocidentais). Para resolver esse problema, eles propuseram uma abordagem nova, construindo vetores de caracteres e vetores de frases e os fundiram antes de enviá-los ao modelo BiLSTM para treinamento. Para realizar os experimentos, foi construído um dataset anotado manualmente, a partir de vários documentos judiciais, como processos criminais, civis e administrativos, obtidos da Rede de Documentos Judiciais Chineses.  Os tipos de entidade  anotados  incluem  nomes  de  pessoas,  organizações,  crimes,  leis  e  regulamentos e penalidades. No total, foram feitas 40.737 anotações entre as diversas EN. No artigo, não há menção à metodologia adotada para a anotação do corpus gerado. 

Até o presente momento, são pouquíssimos os conjuntos de dados, com padrão ouro, de entidades nomeadas do domínio jurídico em português. Os trabalhos desenvolvidos são fragmentados, com poucas classes específicas e de tamanho limitado, o que é um obstáculo para o desenvolvimento de classificadores REN baseados em dados. No trabalho pioneiro realizado por \cite{de2018lener}, os autores disponibilizaram um conjunto de dados de entidades nomeadas, chamado de LeNER-Br, construído a partir de anotações manuais de 66 documentos jurídicos de diversos tribunais brasileiros, entre eles o Supremo Tribunal Federal, Superior Tribunal de Justiça, Tribunal de Justiça de Minas Gerais e Tribunal de Contas da União. Adicionalmente, foram incluídos quatro documentos legislativos, como a Lei Maria da Penha, totalizando 70 documentos anotados. As entidades categorizadas foram "ORGANIZACAO" para organizações, “PESSOA” para pessoas, “TEMPO” para entidades temporárias, “LOCAL” para localizações, "LEGISLACAO" para leis e "JURISPRUDENCIA" para decisões sobre processos judiciais. Ao todo, foram feitas 12.248 anotações de EN. O trabalho não cita as atividades realizadas durante o processo de anotação e nem se foram realizadas avaliações de concordância entre anotadores, porém relata que obteve F1-Score geral de 92,53\%. Para as entidades específicas do domínio jurídico, obteve F1-Scores de 97,00\% e 88,82\% para entidades de Legislação e Jurisprudência, respectivamente.

O estudo realizado por \cite{oliveiraulyssesner}, desenvolveu um conjunto
de dados de entidades nomeadas, chamado de UlyssesNER-Br, a partir de 154 projetos de lei e 800 consultas legislativas da Câmara dos Deputados do Brasil, contendo dezoito tipos de entidades estruturadas em sete classes ou categorias semânticas. Baseadas no HAREM \cite{santos2006golden} foram definidas 5 classes típicas: pessoa, localização, organização, evento e data. Além dessas, foram definidas duas classes semânticas específicas para o domínio legislativo: fundamento do direito e produto do direito. A categoria de fundamentos do direito faz referência a entidades relacionadas a leis, resoluções, decretos, bem como a entidades de domínio específico, como projetos de lei, que são propostas de lei em discussão no parlamento, e consultas legislativas, também conhecidas como solicitações de trabalho feitas pelo parlamentares. A entidade produto da lei refere-se a sistemas, programas e outros produtos criados a partir da legislação.
Os autores relatam que o processo de anotação ocorreu em três etapas. A primeira etapa foi usada como treinamento prático dos anotadores. Nas duas demais etapas, as anotações foram avaliados quanto a concordância entre anotadores usando a medida Kappa de Cohen. Ao final do processo de anotação, as equipes alcançaram a média geral no kappa de Cohen de 90\%. Para as anotações foi usada a ferramenta Inception \cite{klie2018inception}. Não há detalhes da quantidade de anotações anotadadas.
Os modelos de aprendizado de máquina Hidden Markov Model (HMM) e Conditional Random Fields (CRF) foram usados para avaliar o corpus. Os resultados mostraram que o modelo CRF teve melhor desempenho na tarefas de NER, com pontuação média de F1-score de 80,8\% na análise por categorias e 81,04\% na análise por tipos.

\section{Metodologia de Construção da CDJUR-BR}
\label{sec:metolodia-contrucao-CDJUR-BR}

\begin{figure}[h]
\centering
\includegraphics[width=\textwidth]{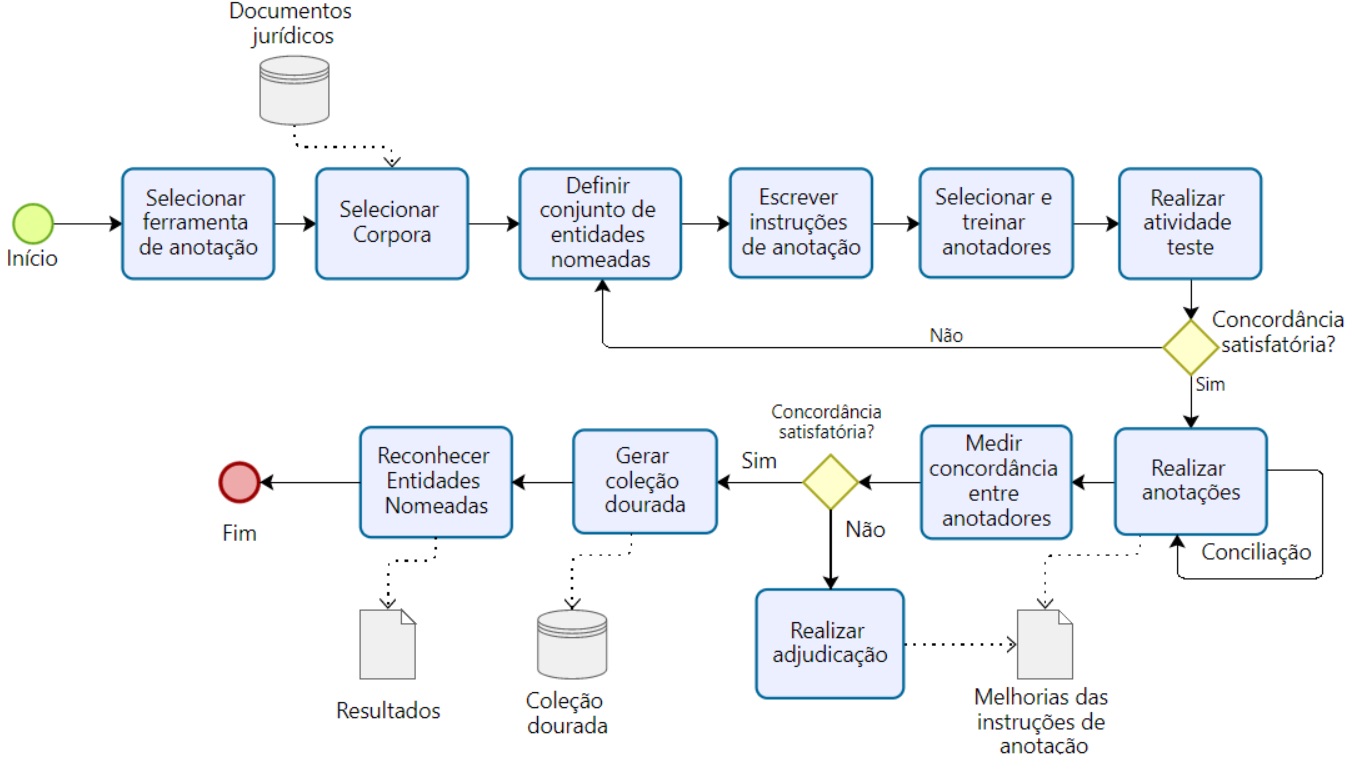}
\caption{Fluxo geral de anotação de corpus.}
\label{fig:fluxo}
\end{figure}

Nesta seção, descreve-se a metodologia proposta para geração de uma Coleção Dourada de entidades do domínio jurídico. As etapas são ilustradas na Figura \ref{fig:fluxo}, consistindo de seleção dos documentos que compõem o \textit{corpus}, definição das entidades a serem anotadas, seleção e treinamento dos anotadores, definição dos critérios de concordância entre anotadores, pre-teste, anotação e, por fim, avaliação e refinamento da anotação. 

Um comitê com três professores da área do Direito e dois da Computação foi formado com o objetivo de definir, juntamente com os especialistas do domínio, os principais parâmetros da CDJUR-BR bem como zelar pela adequada aplicação da metodologia.
\subsection{Seleção da Ferramenta de Anotação}
\label{seleção-ferramenta-anotação}
Inicialmente, foi realizada uma pesquisa de mercado com a finalidade de captar ferramentas de anotação existentes para análise, levando em consideração um conjunto de critérios que indicasse se a ferramenta poderia ser aplicada para o cenário de anotação deste projeto. O conjunto de ferramentas de anotação inicial é composto pelas ferramentas Inception/Webanno, Annotation Lab, Sinapses (CNJ), Doccano, Brat, LightTag, Label Studio, Labelbox, Tagtog, Superannotate, Telus International/Playment, CVAT, Sloth e Dataturk. Em seguida, foi especificado um conjunto de critérios para avaliar quais dessas ferramentas seriam mais adequadas. A lista de critérios que foram analisados são:
\begin{itemize}
    \item Tipo de Dado: quais tipos de dados a ferramenta possibilita anotar (texto, imagem, vídeo, áudio...).
    \item Perfis de Usuário: quais perfis de usuário é possível criar nesta ferramenta (anotador, curador, administrador...).
    \item Fases do Processo de Anotação que Controla: diferentes fases que a ferramenta controla (distribuição dos documentos, curadoria, avaliação de concordância entre anotadores, importação e exportação dos documentos, etc.).
    \item Anotação Automática: se é possível realizar anotação automática de documentos.
    \item Formato do arquivo de saída: quais os formatos de arquivo que é possível exportar as anotações e documentos.
    \item Permite colaboração: se é possível utilizar a ferramenta de uma forma colaborativa.
    \item Custo e suporte técnico: Avaliar o custo de licença para uso da ferramenta e/ou de disponibilidade de suporte técnico.
\end{itemize}
Ao final da avaliação, foram selecionadas e consideradas adequadas às necessidades do projeto as ferramentas Annotation  Lab\footnote{https://www.johnsnowlabs.com/annotation-lab/}, Inception \cite{klie2018inception} e Tagtog \cite{cejuela2014tagtog}. Adotamos a Tagtog por facilidade de iniciação das atividades e disponibilidade de suporte técnico.

\subsection{Seleção dos Documentos do \textit{Corpus}}
\label{seleção-documentos-corpus}
Nessa etapa, o foco é a seleção dos documentos representativos do domínio em questão e baseado em critérios os mais objetivos possíveis \cite{manning1999foundations}. Os critérios definidos foram representatividade e qualidade. Para avaliar a representatividade foram verificadas as classes de processos judiciais de maior ocorrência no Tribunal de Justiça do Estado do Ceará, segundo a tabela de Classes do Conselho Nacional de Justiça do Brasil \cite{silva2013transparencia}: Procedimento Comum Cível, Procedimento do Juizado Especial Cível, Execução Fiscal, Execução de Título Extrajudicial, Inquérito Policial, Ação Penal - Procedimento Ordinário e outras\footnote{Outras Classes compreendem: Relaxamento de Prisão, Execução da Pena, Alimentos, Medidas Protetivas de Urgência Criminal, Busca e Apreensão em Alienação Fiduciária.}. Estas classes representam, conjuntamente, em torno de 85\% dos processos encerrados em 2019 no TJCE. A quantidade de documento por classe foi definida de forma proporcional a frequencia real de processos de 2019 em cada classe, com exceção para a classe Ação Penal que, devido a relevância que a mesma possui, teve um peso maior do que as outras classes. Para o critério qualidade, foram definidos os documentos com 80\% ou mais de palavras válidas da língua portuguesa e com mais de 50 tokens.

O critério de escolha dos documentos que compuseram o corpus, foi a relevância e representatividade, em termos de conteúdo, dos documentos em um processo judicial, determinadas por uma equipe de especialistas do domínio jurídico. Foram selecionados os seguintes documentos: Petição Inicial, Petição, Denúncia, Inquérito, Decisão, Sentença, Despacho e Alegações Finais. A seleção final foi randômica dentre um conjunto de 80 mil documentos dos arquivos do TJCE que atenderam aos critérios acima explanados. A tabela \ref{tabela-composicao-corpus} apresenta aquantidade de documentos que compõem o \textit{corpus} a ser anotado para a CDJUR-BR, por tipo de documento e classe, totalizando 1.216 documentos.

\begin{table}[h!]
\centering
\caption{Composição do \textit{Corpus}.}
\label{tabela-composicao-corpus}
\begin{tabular}{ |c|c|c|c|c|c|c|c| }
\hline
\textbf{Tipo de} &  \textbf{Proc.} & \textbf{Juizado} & \textbf{Execução} & \textbf{Execução} & \textbf{Inquérito} & \textbf{Ação} & \textbf{Outras}\\
\textbf{Documento} &  \textbf{Comum} & \textbf{Especial} & \textbf{Fiscal} & \textbf{Extra-} & \textbf{Policial} & \textbf{Penal} & \textbf{}\\
\textbf{} &  \textbf{Cível} & \textbf{Cível} & \textbf{} & \textbf{judicial} & \textbf{} & \textbf{} & \textbf{}\\
\hline
\hline
Petição Inicial & 35 & 30 & 35 & 33 & 0 & 16 & 29\\
\hline
Petição  & 20 & 18 & 20 & 19 & 0 & 33 & 57\\
\hline
Denúncia & 0 & 0 & 0 & 0 & 12 & 85 & 35\\
\hline
Inquérito & 0 & 0 & 0 & 0 & 53 & 54 & 31\\
\hline
Decisão & 21 & 21 & 20 & 0 & 22 & 71 & 32\\
\hline
Sentença & 20 & 20 & 18 & 20 & 22 & 30 & 48\\
\hline
Despacho & 11 & 12 & 13 & 30 & 2 & 6 & 22\\
\hline
Alegações Finais & 0 & 0 & 0 & 0 & 2 & 115 & 23\\
\hline
\textbf{Total por Classe} & 107 & 101 & 106 & 102 & 113 & 410 & 277\\ 
\hline
\hline
\textbf{Total TJCE} & & & & & & & 1.216\\
\hline
\end{tabular}
\end{table}

\subsection{Definição das Categorias de Entidades Nomeadas}
\label{sec:definição-categorias-entidades-nomeadas}

As  categorias  de  entidades  nomeadas  mais  comuns  são  pessoas,  organizações, normas e  localizações \cite{mikheev1999named}.  Porém,  neste  trabalho,  elas foram  definidas para representar mais detalhadamente as entidades específicas  do  domínio  jurídico. Essas entidades foram escolhidos por representantes do TJCE e do MPCE visando a posterior automatização de tarefas de sumarização de textos, similaridade de processos, classificação de assunto, sugestão de penas e consulta de jurisprudência. Por exemplo, a Categoria \textit{Pessoa} foi especificada em 9 entidades que normalmente estão presentes em um processo judicial, quais sejam: autor, advogado, réu, vítima, testemunha, juiz, promotor, autoridade policial e outras. Os \textit{Endereços} foram especificados em 6 entidades para identificar os diversos endereços presente em um processo judicial. A categoria \textit{Norma} foi especificada em três entidades. A primeira descreve as referências legais diretamente relacionadas ao assunto principal do processo. A segunda refere-se às normas que contextualizam o documento jurídico em questão (e.g. petições, decisões, etc.). A terceira especifica as decisões de jurisprudências mencionadas nos processos. De forma similar, foram feitas especificações para i)\textit{Prova} especificando as provas usadas pela acusação e defesa durante o processo; ii) \textit{Pena} para  identificar  as  sanções  aplicadas; iii) \textit{Sentença} para  identificar  as  sentenças  proferidas  pelos  juízes; A relação completa apresentada de entidades com a respectiva quantidade de entidades rotuladas está na tabela \ref{estatistica-entidades}.

\subsection{Instruções de Anotações}
\label{sec:instrucoes-anotações}
Uma vez definidas as entidades nomeadas a serem anotadas, foi iniciada a elaboração das instruções de anotação. As instruções são diretrizes que deverão ser seguidas pelos anotadores com o propósito de alcançar maior concordância nas anotações realizadas. Quanto maior a concordância, pressupõe-se maior a qualidade da coleção dourada que será criada. Nesse sentido, as  instruções foram sendo aprimoradas continuamente, ao longo do trabalho por meio de reuniões entre as equipes de anotadores e o comitê técnico e de gestão dos trabalhos. Como resultado dessas reuniões, as dúvidas dos anotadores eram esclarecidas e exemplos mais detalhados foram compartilhados em novas versões das  instruções para evitar mal-entendidos. 

\subsection{Seleção e Treinamento dos Anotadores}
\label{sec:seleção-treinamento-anotadores}

No processo de anotação, segundo \cite{hovy2010towards}, a abordagem geral adotada é se usar anotadores que são razoavelmente semelhantes em educação e cultura, realizar treinamento e disponibilizar um  manual  bastante  específico  para  que  se  consiga  a boa  correspondência  nas anotações. 
Neste  trabalho,  foram  formadas  três  equipes  de especialistas no domínio jurídico, selecionados entre  os colaboradores  dos  parceiros  do  projeto para realizar as anotações: Uma equipe composta, inicialmente, por 14 juízes estaduais, de ambos os sexos, com experiência entre 5 e 15 anos, realizou anotações em documentos das classes CNJ: Procedimento Comum Cível, Juizado Especial Cível, Execução Fiscal, Execução Extrajudicial, Ação Penal e outras. Uma outra equipe composta por 19 promotores de justiça e técnicos do judiciário, de ambos os sexos, com experiência entre 5 e 10 anos, realizaram anotações em documentos das classes CNJ: Procedimento Comum Cível, Relaxamento de Prisão e Execução da Pena. A terceira equipe foi formada por 3 professores de direito, com nível de mestrado ou doutorado e experiência na atividade jurídica profissional superior a 10 anos. Esta equipe trabalhou nas anotações dos mesmos documentos da equipe de juízes e nas revisões da fase de adjudicação.
Concluídas as seleções, os anotadores passaram por treinamento visando aumentar a concordância entre os anotadores. O treinamento teve duração de 90 minutos e visava suprir os especialistas de habilidades relacionadas diretamente ao processo de anotação (instruções de anotação), uma vez que estes possuíam sólidos conhecimentos dos procedimentos e linguagem  jurídica presente nos documentos. Foram abordados os seguintes tópicos: Contextualização do projeto e finalidade da rotulação de documentos; Conceitos básicos de aprendizado automático supervisionado; Resumo da metodologia do processo de anotação; Descrição das  entidades a serem rotuladas e treinamento prático do software Tagtog.
\subsection{Atividade Teste}
\label{sec:atividade-teste}
Após o treinamento, foi realizada uma atividade teste para que os anotadores pudessem praticar as instruções de anotação e ganhar familiaridade com o software \textit{Tagtog} e com o processo em si. Durante essa fase, os anotadores puderam anotar algum fragmento do \textit{corpus} de treino, de forma a determinar a viabilidade da metodologia e das instruções de anotação. Ao final, desta etapa, houve uma revisão das instruções de anotação e se chegou a configuração final das entidades a serem rotuladas. A atividade teste se mostrou de grande importância para a execução dos trabalhos, pois foi um momento de grande interação entre os anotadores,  que puderam esclarecer dúvidas sobre questões jurídicas, como também puderam entender melhor o objetivo das anotações com os esclarecimentos obtidos com os especialistas em PLN. 

\subsection{Processo de Anotação do \textit{Corpus}}
\label{processo-anotação-corpus}
\begin{figure}[h]
\centering
\includegraphics[width=\textwidth]{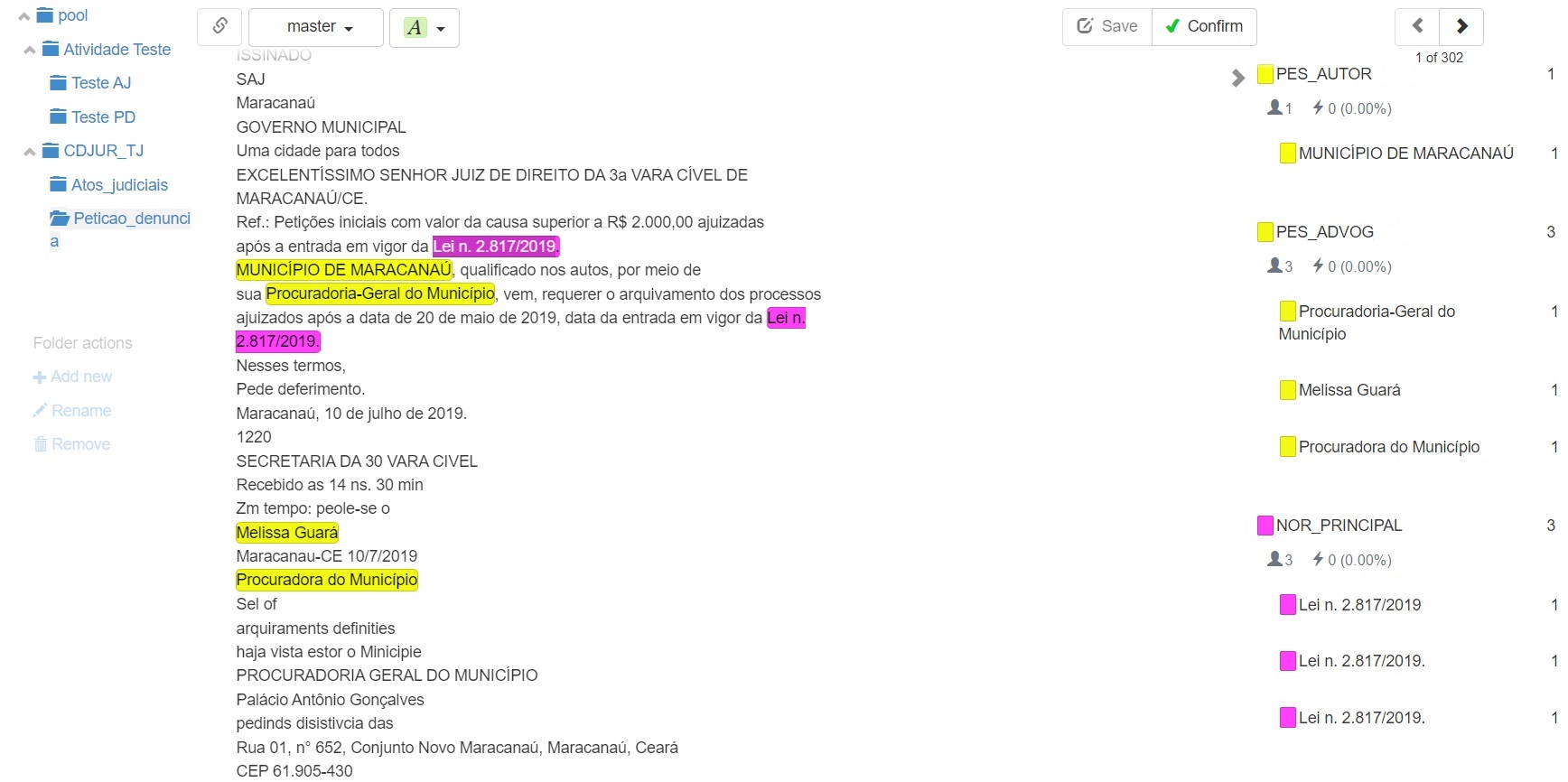}
\caption{Exemplo de anotação de documento.}
\label{fig:anotacao}
\end{figure}

A figura \ref{fig:anotacao} exemplifica a anotação de um documento usando \textit{Tagtog}. Na figura pode-se ver o realce de algumas entidades que foram anotadas.

Cada documento foi anotado por dois anotadores diferentes. Os anotadores receberam o manual de instruções e, cada um deles, teve liberdade para realizar seu trabalho da ordem que preferisse. Alguns optaram por realizar as anotações seguindo o texto de cima a baixo e identificando as diferentes entidades que reconhecia. Outros, ao identificar uma determinada entidade, percorriam o texto inteiro em busca de ocorrências semelhantes e, só depois, retomavam ao início para identificar uma nova entidade e repetir o processo novamente até que completasse o trabalho em um documento.

A primeira etapa do processo de anotação ocorreu em 2,5 meses, quando todos os documentos haviam sido anotados por pelo menos dois anotadores. Depois dessa etapa, a CDJUR-BR entrou no ciclo de avaliação e refinamento.
\subsection{Avaliação da Concordância entre Anotadores}
\label{sec:avaliação-concordância-entre-anotadores}
Para garantir maior confiabilidade das anotações, foi utilizado o coeficiente Kappa de Cohen \cite{mchugh2012interrater} para avaliar a concordância entre anotadores. A Tabela \ref{kappa}, mostra os resultados obtidos para cada categoria de entidades, para os documentos que alcançaram o coeficiente Kappa superior a 0,50 após o processo de anotação. A categoria Pessoa obteve Kappa 0,79, indicando a maior concordância entre os anotadores. O resultado é muito bom, pois esta categoria é composta por 9 outras entidades mais específicas (conforme tabela \ref{estatistica-entidades}) e isso poderia ter levado os anotadores a dúvidas entre as diversas partes representadas em um processo. Na categoria Prova se percebe que houve maior dificuldade de consenso na tipificação da prova. O que refletiu na categoria com o mais baixo coeficiente Kappa (0,46). Importante destacar que estes resultados foram os obtidos antes das revisões de um terceiro anotador (processo de adjudicação).
\begin{table}[h!]
\centering
\caption{Resultados das Anotações por Categorias}\label{kappa}
\begin{tabular}{ |c|c|c| }
\hline
\textbf{Categoria} &  \textbf{Anotações} & \textbf{Kappa}\\
\hline
Pessoa & 15.149 & 0.79\\
Prova  & 1.696 & 0.46\\
Pena & 205 & 0.64\\
Endereço & 2.041 & 0.73\\
Sentença & 106 & 0.75\\
Norma & 6.216 & 0.76\\
\hline
\textbf{Total} &  \textbf{25.413} & \textbf{0,69}\\
\hline
\end{tabular}
\end{table}
\subsection{Processo de Adjudicação}
\label{sec:processo-adjudicação}
 
Ao final da fase de anotações, 732 documentos obtiveram coeficiente Kappa médio de 0,69. Porém, como em qualquer qualquer tarefa complexa, o acordo total nunca é possível, mesmo após as reconciliações. Então, foi adotada a estratégia de adjudicação, que consiste em um terceiro especialista revisar os casos de desacordo e decidir \cite{hovy2010towards}. Neste trabalho, o processo de adjudicação foi feito com os documentos que o Kappa foi inferior a 0,50. Por este critério, 166 documentos foram selecionados. As revisões consistiram em um terceiro revisor receber as anotações dos dois anotadores iniciais (uma união das anotações), então o revisor poderia decidir por acatar uma das anotações ou mesmo fazer uma nova anotação diferente das duas anteriores. Além disso, no decorrer das anotações, alguns dos especialistas selecionados deixaram os trabalhos, com isso, 176 documentos remanescentes foram anotados pela equipe de adjudicação. 
\subsection{Refinamento das Normas Legais}
\label{avaliação-refinamento-normas-legais}
Data a importância da categoria normas para os objetivos deste trabalho e para o domínio jurídico, a avaliação da qualidade das entidades que descreviam normas mereceu atenção especial da equipe. Após os primeiros experimentos realizadas com os sistemas REN desenvolvidos, específicamente, para a CDJUR-BR, verificou-se que havia muitas entidades reconhecidas pelos sistemas que não foram identificadas pelos anotadores e, portanto, não constavam na coleção dourada. Diante dessa constatação, a equipe de especialista revisores recebeu um relatório com as normas identificadas pelo REN e analisaram se eram, de fato, normas e a que tipo específica deveriam ser associadas, ou seja, se seriam uma norma principal, acessória ou uma jurisprudência. Esta etapa de melhoria e refinamento da CDJUR-BR, propiciou uma ampla revisão das diretrizes de caráter jurídico das entidades, como também, quanto aos critérios relacionados a limites das anotações. Ao final dessa etapa, foram adicionadas 4.338 novas entidades de normas jurídicas. 

\begin{figure}[h]
\centering
\includegraphics[width=\textwidth]{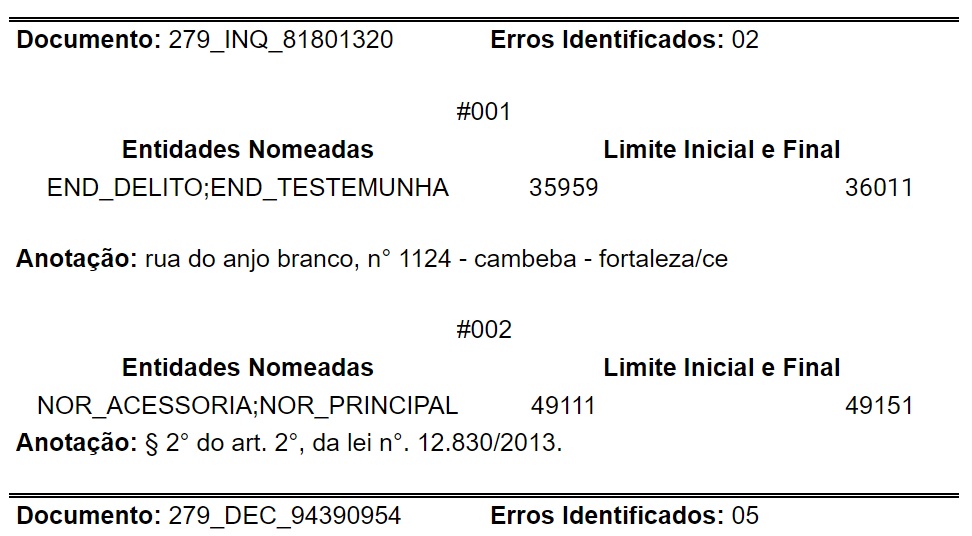}
\caption{Exemplo do Relatório de Revisão.}
\label{fig:relatorio-revisao}
\end{figure}

Adicionalmente, foi desenvolvido um aplicativo em Python para realizar a correção automática de documentos em que o erro se limitasse a discordância quanto aos limites inicial e final da anotação. Para isso, a equipe gestora definiu que fosse feita a união das sentenças marcadas. Quando o aplicativo encontrava discordância entre entidades anotadas, gerava um relatório apontando os documentos e as entidades que apresentavam as discordâncias. Com isso, os revisores puderam realizar suas atividades corretivas com maior praticidade, além de terem a liberdade de realizar novas anotações se assim julgassem necessárias. A figura \ref{fig:relatorio-revisao} exemplifica o relatório de revisão usado. No exemplo, o documento apresenta dois erros: O primeiro é uma discordância entre as entidades END\_DELITO e END\_TESTEMUNHA. O limite inicial da marcação é na posição 35.959 e a posição final é a 36.011, portanto a marcação tem 52 caracteres (incluindo espaços). Diante dessa informação o revisor analisaria o documento para decidir qual EN era a correta. O segundo erro é entre as entidades NOR\_ACESSORIA e NOR\_PRINCIPAL, com início em 49.111 e limite final em 49.151 (comprimento de 40 caracteres). O início e fim da anotação foi incluida para dar uma noção de onde se encontrava a anotação no documento.

A coleção dourada final contém 1.074 documentos. Ao todo foram 44,526 entidades rotuladas. A tabela \ref{estatistica-entidades} apresenta estatísticas do \textit{corpus} anotado.

\begin{table}[h]
\centering
\caption{Estatística do corpus anotado.}\label{estatistica-entidades}
\begin{tabular}{ |c|c|c|c|c|c| }
\hline
\textbf{Categoria} &  \textbf{Anotações} & \textbf{\%} & \textbf{Entidade Nomeada} & \textbf{Anotações} & \textbf{\%}\\
\hline
\hline
Pessoa & 24.844 & 55,80 & PES-ADVOG & 735 & 1,65 \\
        &  &  & PES-AUTOR & 1.259 & 2,83 \\
        &  &  & PES-AUTORID-POLICIAL & 2.012 & 4,52 \\
        &  &  & PES-JUIZ & 576 & 1,29 \\
        &  &  & PES-OUTROS & 6.003 & 13,48 \\
        &  &  & PES-PROMOTOR-MP & 363 & 0,82 \\
        &  &  & PES-REU & 8.773 & 19,70 \\
        &  &  & PES-TESTEMUNHA & 2.967 & 6,66 \\
        &  &  & PES-VITIMA & 2.156 & 4,84 \\
\hline
Prova  & 3.318 & 7,45 & PROVA & 3318 & 7,45 \\
\hline
Pena & 407 & 0.91 & PENA & 407 & 0,91 \\
\hline
Endereço & 2.065 & 4,64 & END-AUTOR & 132 & 0,30 \\
  &   &   & END-DELITO & 466 & 1,05 \\
  &   &   & END-OUTROS & 355 & 0,80 \\
  &   &   & END-REU & 693 & 1,56 \\
  &   &   & END-TESTEMUNHA & 295 & 0,66 \\
  &   &   & END-VITIMA & 124 & 0,28 \\
\hline
Sentença & 172 & 0.39 & SENTENÇA & 172 & 0,39\\
\hline
Norma & 13.720 & 30,81 & NOR-ACESSORIA & 5.767 & 12,95 \\
  &   &   & NOR-JURISPRUDÊNCIA & 1.823 & 4,09 \\
  &   &   & NOR-PRINCIPAL & 6.130 & 13,77 \\
\hline
\textbf{Total} & 44.526 & 100 &  \textbf{Total} & 44.526 & 100 \\
\hline
\end{tabular}
\end{table}
\section{Resultados}
\label{chap:resultados}

Entre os objetivos deste trabalho está a criação da Coleção Dourada de Entidades Nomeadas da Justiça Brasileira (CDJUR-BR). Este objetivo teve como requisito fundamental a confiabilidade e consistência do \textit{corpus} criado para assegurar sua viabilidade no treinamento de algoritmos de aprendizado de máquina utilizado nas soluções de \textit{Legal AI}. Dada a importância estratégica destes requisitos, nossos esforços foram balizados pelas Questões de Pesquisa definidas na subseção \ref{sec:problemática-questoes-pesquisa}. A \textit{QP1} (Como elaborar uma metodologia de anotações manuais de entidades nomeadas que contemple as especificidades e complexidades do domínio jurídico?) foi respondida nas etapas de  desenvolvimento das anotações, descritas na metodologia aqui apresentada (seção \ref{sec:metolodia-contrucao-CDJUR-BR}) e validada pelas avaliações de concordância entre anotadores (alcançando o coeficiente de Kappa geral de 0,69) e por meio das etapas de conciliação, adjudicação e atividades extras de refinamento (subseções \ref{sec:avaliação-concordância-entre-anotadores}, \ref{sec:processo-adjudicação} e \ref{avaliação-refinamento-normas-legais}) que possibilitaram a adição de 19.113 anotações a coleção dourada final.

Para responder a \textit{QP2} (A coleção dourada gerada é adequada para o treinamento e validação de modelos de \textit{Legal AI}?), avaliamos a CDJUR-BR na tarefa de Reconhecimento de Entidades Nomeadas (REN) através de alguns cenários de experimentos que descrevemos a seguir.

\subsection{Avaliação da CDJUR-BR na Tarefa de REN}

\subsubsection{Cenários de Experimentos}
\label{sec:cenarios_experimentos}

Para avaliar a CDJUR-BR, realizamos vários experimentos em cenários que diferem quanto aos conjuntos de dados utilizados para treino ou para teste e quanto as estratégias para representar as entidades. Em todos os cenários, utilizamos os modelos descritos na seção \ref{sec:modelos-NER} e as métricas Precision, Recall e F1-Score nas avaliações. Como há uma grande diferença entre as quantidades de anotações por categoria de entidades nomeadas, foi desenvolvida uma heurística para manter a mesma propoporção de exemplos da coleção completa quando dividimos em conjuntos de treino, validação e teste. Dessa forma, evitou-se que os conjuntos de validação e teste ficassem com poucos exemplos, especialmente nas categorias Pena e Sentença. Os conjuntos de treino, validação e teste ficaram com 68,07\%, 15,21\% e 16,72\% das amostras, respectivamente.
\begin{itemize}
    \item \textbf{Cenário 1 - C1. Reconhecimento das entidades específicas da CDJUR-BR.} Neste cenário, utilizamos os dados da CDJUR-BR para treinar os modelos para o reconhecimento das entidades específicas definidas na CDJUR-BR. Nosso propósito, com este cenário, é demonstrar a viabilidade da CDJUR-BR para o treinamento de modelos REN no domínio jurídico em língua portuguesa.
    \item \textbf{Cenário 2 - C2. Reconhecimento das categorias da CDJUR-BR.} Agrupamos as entidades específicas da CDJUR-BR nas seguintes categorias: Pessoa (categoria formada por todas as entidades específicas que se referem a Pessoa, ou seja, todos os tokens representando Pessoa foram unicamente etiquetados como Pessoa), Legislação (categoria formada pelas entidades NOR-ACESSÓRIA e NOR-PRINCIPAL), Jurisprudência (categoria formada pela entidade NOR-JURISPRUDÊNCIA) e Local (categoria formada por todas as EN específicas que se referem a Endereço). Em resumo, nesse cenário treinamos e avaliamos o modelo com a CDJUR-BR, porém, o reconhecimento é em nível de categorias. Com isso, esse cenário nos possibilitará fazer comparações com a coleção LENER-BR.
    \item \textbf{Cenário 3 - C3. Reconhecimento das categorias de entidades da LENER-BR a partir de modelo treinado com LENER-BR.} Neste cenário, os modelos REN  foram treinados utilizando o conjunto de dados de treino do LENER-BR, que contém 6 diferentes EN: Pessoa, Jurisprudência, Tempo, Local, Legislação e Organização. Esse canário nos apresentará o desempenho que os modelos alcançarão com a LENER-BR para termos como referência comparatica de desempenho do REN.
    \item \textbf{Cenário 4 - C4. Reconhecimento das entidades do LENER-BR a partir de modelo treinado com CDJUR-BR.} Neste cenário, agrupamos as entidades específicas do conjunto de treino da CDJUR-BR nas categorias Pessoa, Legislação, Jurisprudência e Local e treinamos os modelos REN. Na fase de teste, avaliamos para o reconhecimento das entidades Pessoa, Legislação, Jurisprudência e Local com os dados  do LENER-BR. Esse cenário nos possibilitará avaliar a capacidade de generalização do modelo REN treinado com a CDJUR-BR quando usado com outros documentos (no caso, os documentos que compuseram a LENER-BR). 
    \item \textbf{Cenário 5 - C5. Reconhecimento das categorias de entidades da CDJUR-BR a partir de modelo treinado com LENER-BR.} Nesse cenário, os modelos REN foram treinados utilizando o conjunto de dados de treino do LENER-BR. Porém, na fase de teste, avaliamos os modelos no reconhecimento das seguintes categorias da CDJUR-BR: Pessoa, Legislação (categoria formada pelas entidades NOR-ACESSÓRIA e NOR-PRINCIPAL), Jurisprudência (categoria formada pela entidade NOR-JURISPRUDÊNCIA) e Local (categoria formada por todas as EN específicas que se referem a Endereço). Esse cenário buscará demonstrar quão capaz serão os modelos treinados com a LENER-BR em reconhecer entidades de outro \textit{corpus} (no caso, os documentos que compuseram a CDJUR-BR). Adicionalmente, os resultados nos permitirão comparar a capacidade de generalização dos modelos REN treinados com a CDJUR-BR e LENER-BR (ao se comparar os resultados obtidos no C4 com os resultados obtidos no C5).  
\end{itemize}

\subsection{Modelos para o NER}
\label{sec:modelos-NER}
Para realizar os experimentos, desenvolvemos três modelos de aprendizagem automática para o reconhecimento de entidades nomeadas (REN). Utilizamos o modelo com o SPACY para estabelecer uma linha de base para os nossos experimentos. O SPACY é amplamente utilizado, não requer muito conhecimento e tempo para ser construído e consegue resultados razoáveis na tarefa de REN. Os outros dois modelos, BI-LSTM + CRF e o BERT conseguem alcançar o desempenho estado-da-arte, por isso os escolhemos visando verificar qual deles tirará melhor proveito com os dados disponíveis. Seguem os detalhes de cada modelo: 

\begin{itemize}
    \item \textbf{Bidirectional Long Short-Term Memory (BI-LSTM) acrescido de uma Camada CRF (BI-LSTM + CRF)} \cite{graves2005framewise,hochreiter1997long,lafferty2001conditional}. A entrada do modelo é uma sequência de representações vetoriais de palavras individuais construídas a partir da concatenação de embeddings de palavras e embeddings de nível de caractere. Para a tabela de pesquisa de palavras, usamos o GloVe [15] (vetor de palavras pré-treinadas em um corpus multi-gênero formado por textos em português do Brasil e da Europa [16]). Usamos 10 épocas e lotes com 10 amostras de tamanho; usamos o otimizador SGD com uma taxa de aprendizado de 0,015.
    \item \textbf{Bidirectional Encoder Representations from Transformers (BERT)} \cite{devlin2018bert}. Utilizamos a abordagem baseada em ajuste fino com o modelo pré-treinado BERTimbau \cite{souza2020bertimbau}, 10 épocas e tamanho de lote de 8 amostras. Como otimizador usamos o ADAM com uma taxa de aprendizado de \textit{$1x10^{-5}$}.
    \item \textbf{SPACY}. Treinamos o modelo para o componente NER do pipeline do SPACY \cite{spacy2}, iniciando a partir do pacote em Português \textit{pt\_core\_news\_sm}.
\end{itemize}

\subsection{Resultados e Discussões}
\label{sec:resultados-dicussões}

A tabela \ref{resultados c1} apresenta os resultados obtidos no conjunto de teste para o cenário 1 (C1). O modelo com o BERT, comparativamente, obteve o melhor desempenho na grande maioria das entidades, alcançando um F1-Score médio macro de 0,58. E, as entidades PES-AUTORID-POLICIAL (0,90), NOR-JURISPRUDÊNCIA (0,89), PES-PROMOTOR-MP (0,88), NOR-ACESSÓRIA (0,82) obtiveram os mais altos F1-Scores. Esses resultados comprovam a viabilidade da CDJUR-BR para ser usada em modelos de aprendizado de máquina em soluções de \textit{Legal AI (QP2)}

O reconhecimento de entidades nomeadas com os modelos LSTM + CRF e SPACY obtiveram desempenhos inferiores nos experimentos realizados, alcançando um F1-Score médio macro de 0,55 e 0,42, respectivamente. Apesar do modelo LSTM + CRF ter obtido um desempenho um pouco inferior ao BERT, as entidades NOR-JURISPRUDÊNCIA (0,90), PENA (0,56), PES-AUTOR (0,59), PES-JUIZ (0,79) e PROVA (0,47) obtiveram resultados iguais ou melhores de F1-Score dentre os modelos avaliados. Já o modelo implementado com o SPACY, obteve resultado igual para PES-AUTOR (0,59) e foi melhor para a entidade SENTENÇA (0,29). Ao realizar uma análise sintética dos resultados obtidos pelo melhor modelo (BERT), percebe-se que grande parte dos erros são de predições de tokens do tipo "O" (formato IOB, \cite{ramshaw1999text}). Esse tipo de ocorrência correspondeu a quase 30\% dos erros verificados. Nas entidades que formam categorias, também, observa-se erros por ambiguidade de entidades da mesma categoria: Na categoria Endereço os erros de ambiguidade chegam a 16\%, em Normas 8\% e em Pessoa os erros de ambiguidade na mesma categoria alcançam 16\%. A seguir, analisamos o melhor resultado obtido por cada EN no Cenário 1. Em seguida, avaliamos os demais cenários (C2,C3, C4 e C5).
\begin{table}[h]
\centering
\caption{Resultados de F1-score para o Reconhecimento das Entidades Específicas (C1) utilizando os modelos BI-LSTM+CRF, SPACY e BERT}
\label{resultados c1}
\begin{center}
\begin{tabular}{| c || c || c || c || c |}
\hline
\textbf{Entidade Nomeada} & {\textbf{BI-LSTM+CRF}} & {\textbf{SPACY}}& {\textbf{BERT}} & {\textbf{Suporte}}\\ 
\hline
\hline
END-AUTOR &\textbf{0.56}&0.31&0.33&18\\ 
\hline
END-DELITO &0.72&0.45&\textbf{0.73}&61\\ 
\hline
END-OUTROS &0.00&0.02&0.16&81\\ 
\hline
END-REU &0.55&0.59&\textbf{0.71}&152\\ 
\hline
END-TESTEMUNHA &0.27&0.26&\textbf{0.67}&68\\ 
\hline
END-VITIMA &0.06&0.00&\textbf{0.22}&27\\ 
\hline
NOR-ACESSÓRIA &0.79&0.79&\textbf{0.82}&990\\ 
\hline
NOR-JURISPRUDÊNCIA &\textbf{0.90}&0.87&0.89&333\\ 
\hline
NOR-PRINCIPAL &0.67&0.71&\textbf{0.77}&791\\ 
\hline
PENA &\textbf{0.56}&0.39&0.50&82\\ 
\hline
PES-ADVOG &0.54&0.22&\textbf{0.63}&122\\ 
\hline
PES-AUTOR &\textbf{0.59}&\textbf{0.59}&0.56&169\\ 
\hline
PES-AUTORID-POLICIAL &0.87&0.66&\textbf{0.90}&300\\ 
\hline
PES-JUIZ &\textbf{0.79}&0.50&0.78&83\\ 
\hline
PES-OUTROS &0.54&0.44&\textbf{0.58}&1.210\\
\hline
PES-PROMOTOR-MP &0.81&0.27&\textbf{0.88}&57\\
\hline
PES-REU &0.64&0.57&\textbf{0.71}&1.503\\ 
\hline
PES-TESTEMUNHA &0.57&0.45&\textbf{0.64}&519\\
\hline
PES-VÍTIMA &0.33&0.23&\textbf{0.46}&405\\
\hline
PROVA &\textbf{0.47}&0.29&0.34&461\\
\hline
SENTENÇA &0.00&\textbf{0.29}&0.00&11\\
\hline
\hline
\textbf{F1-micro avg} &0.64&0.55&\textbf{0.67}&7.443\\
\hline
\textbf{F1-macro avg} &0.53&0.42&\textbf{0.58}&7.443\\ 
\hline
\textbf{F1-weighted avg} &0.62&0.54&\textbf{0.67}&7.443\\ 
\hline
\end{tabular}
\end{center}
\end{table}
\subsubsection{Análise dos resultados para o cenário 1.}

O modelo BERT teve melhor desempenho, ainda assim, na análise detalhada se pode constatar que ele teve como maior dificuldade a desambuiguação de entidade do tipo "O". Estes tipos de tokens representaram mais de 60\% das predições. Os FN do tipo "O" representaram mais da metade desse tipo de erro. Segue análise análítica dos resultados das entidades para o C1.\\
\textbf{END-AUTOR:} O modelo BI-LSTM + CRF obteve o melhor desempenho, com F1-Score de 0,56. As predições apresentam uma precisão moderada devido o alto índice de falsos positivos. As predições são confundidas, principalmente, com com tokens não-anotados ("O") ou END-REU. O Recall é razoável (0,56). A pequena quantidade de exemplos pode contribur para a baixa performance do modelo para essa EN.\\
\textbf{END-DELITO:} Esta EN foi a de melhor desempenho na categoria, alcançcando o F1 de 0,73. Esse resultado veio pelo excelente Recall (0,93), porém, a precisão foi moderada (0,59). As predições são confundidas, principalmente, com tokens não-anotados ("O") ou END-REU.\\
\textbf{END-OUTROS:} Esta EN foi a de pior resultado no NER para os endereços. A precisão e o Recall foram muito baixos (0,13 e 0,20, respectivamente), tendo uma altíssima quantidade de falsos positivos (87\%) do tipo "O".\\
\textbf{END-REU:} Esta entidade obteve muita harmonia entre a Precisão (0,70) e o Recall (0,72). A maior quantidade de erros acontece entre as entidades da mesma categoria.\\
\textbf{END-TESTEMUNHA:} Alcança bom Recall (0,71), porém a Precisão é moderada (0,63). As predições erradas são, em sua maioria, entre as entidades de mesma categoria, mas também tokens "O".\\
\textbf{END-VITIMA:} Essa EN teve o desempenho muito baixo (F1 de 0,22). Precisão (0,44) e Recall (0,15) foram baixos, com muitos FP e FN da mesma categoria.\

\textbf{Resumo da Análise para Endereço:} Apesar de algumas EN específicas apresentarem baixo F1, o NER obteve bom desempenho para a categoria Endereço, alcançando F1 de 0,72. Este resultado pode ser explicado por vários erros de predição ocorrerem entre entidades da mesma categoria e, pelo fato das EN de pior desempenho contarem com poucos exemplos (suporte).\\
\\
\textbf{NOR-ACESSÓRIA:} O NER obteve excelente desempenho em reconhecer essa entidade (F1 de 0,82). Este desempenho foi melhor em função do Recall elevado (0,86), tendo a maior parte dos FN apontados na Norma Principal e nos tokens tipo "O". A precisão foi mais baixa (0,79), principalmente, pelos FP do tipo "O" e de Norma Principal.\\
\textbf{NOR-JURISPRUDÊNCIA:} Esta norma é a que o NER, com o modelo BI-LSTM + CRF, obtenhe melhor desempenho (F1 de 0,90). A precisão alcançada de 0,86, mostra erros de FP em tokens sem marcação ("O") e, curiosamente, em PES-OUTROS. O Recall alcançou 0,95, sendo os poucos FN de tokens tipo "O".\\
\textbf{NOR-PRINCIPAL:} Apresenta um bom desempenho (F1 0,77). A Precisão de 0,72 puxou o F1 para baixo. As predições se confundem muito com as outras entidades da mesma categoria mas, principalmente, com os tokens sem anotação ("O"). Quanto ao Recall (0,84), a maioria dos erros estão nas entidades da mesma categoria.\

\textbf{Resumo da Análise para Norma:} Esta categoria foi a que obteve o mais alto F1 (0,90). A Precisão é de 86\%, sendo os erros mais comuns as predições de tokens tipo "O". O Recall alcançou 0,96, com a maioria dos FN de entidades da mesma categoria.\\
\\
\textbf{Pena:} A Precisão do modelo está moderada (0,60), sendo a maior quantidade de erros (FP) de predições de tipo "O". O Recall é ruim, alcançando 0,43, com os tokens "O" indicados como FN. O F1 de 0,56 foi obtido com o modelo BI-LSTM + CRF.\\
\\
\textbf{PES-ADVOG:} A Precisão foi de 58\%, com o modelo confundindo os tokens não-anotados ("O") e com PES-OUTROS. O Recall alcançou 0,69, sendo os tokens "O" a maioria dos FN.\\
\textbf{PES-AUTOR:} As predições desta entidade tiveram muitas ambiguidades com entidades da mesma categoria e FP do tipo "O", levando a Precisão de 49\%. Já o Recall de 0,66 se caracterizou por muitos FN do tipo "O". O F1 de 0,59 foi obtido com os modelo BI-LSTM + CRF e SPACY.\\
\textbf{PES-AUTORID-POLICIAL:} Esta EN é a que o NER obteve melhor desempenho, com F1 de 0,90. A Precisão alcançada de 0,87, mostram erros de FP em tokens sem marcação ("O") e, alguns poucos da mesma categoria. O Recall foi excelente (0,93), sendo a maior parte dos FN da própria categoria.\\
\textbf{PES-JUIZ:} Apresenta um bom desempenho (F1 0,79) com o modelo BI-LSTM + CRF. A Precisão de 0,83 se caracterizou por erros nos tokens "O". O Recall (0,79) reduziu o F1 e, os erros estão nas entidades de mesma categoria e , também, nos tokens sem anotação ("O").\\
\textbf{PES-OUTROS:} Obteve resultados bastante intrigantes, pois a Precisão e o Recall foram, apenas, moderados, apesar dessa EN ter muitos exemplos anotados. A Precisão de 0,60 se caracteriza por muitos FP da mesma categoria, mas principalmente, de tokens tipo "O". O Recall, também, foi mediano, alcançando 0,56, com muitos FN da mesma categoria, do tipo "O" e até em NOR-JURISPRUDÊNCIA.\\
\textbf{PES-PROMOTOR-MP:} Entidade obteve excelente resultado no NER, com F1 de 0,88. A Precisão alcançada de 0,89, mostra poucos erros de FP em tokens sem marcação ("O"). O Recall, também, foi excelente (0,88), com alguns poucos FN da própria categoria e de tokens sem anotação ("O").\\
\textbf{PES-REU:} Esta EN obteve excelente desempenho (F1 0,71) e  harmonia entre a Precisão (0,70) e o Recall (0,72). A maior quantidade de FP é semelhante entre entidades de mesma categoria e tipo "O". Da mesma forma, os FN também são equilibrados entre as entidades de mesma categoria e os tokesn tipo "O".\\
\textbf{PES-TESTEMUNHA:} Com F1 de 0,64, essa entidade teve as muitos erros de predições da mesma cateoria (PES-REU, PES-OUTROS, PES-VITIMA). Mas, também, confundio com os tokens tipo "O". Com isso a Precisão ficou moderada (0,66). O maior quantidade de FN foi semelhante ao ocorrido com a Precisão, porém a maior quantidade de erros ocorreu com os tokens sem anotação, levando o Recall a 0,62.\\
\textbf{PES-VITIMA:} Esta entidade foi a de pior desempenho na categoria (F1 0,46). A maior quantidade de erros de Predição (FP) estão entre as entidades da mesma categoria, porém, há muitos erros com tokens não anotados. Com isso, a Precisão foi, apenas, 045. O Recall foi de 0,47, com muitos FN da mesma categoria, de tokens tipo "O" e alguns da entidade Prova.\

\textbf{Resumo da Análise para Pessoa:} A categoria Pessoa alcançou F1 de 0,81, que é um resultado muito bom. A maior quantidade de FP e FN estão nos tokens tipo "O", porém, há uma quantidade quase equivalentes dos mesmos erros entre entidades da mesma categoria.\\
\\
\textbf{Prova:} O NER não obteve bom desempenho para esta entidade (F1 0,47). A Precisão foi excelente (0,87), com poucos FP, que ocorreram mais nos tokens "O". O Recall ficou baixo, em 0,33, com FN em entidades de Normas, Prova, Pessoa e, principalmente, nos tokens "O". \\
\\
\textbf{Sentença:} O modelo REN com o SPACY foi o único a conseguir identificar essa entidade. Ainda assim, obteve baixíssimo Recall (0,18), sendo a maioria dos FN de tokens tipo "O". O modelo fez poucas predições, mas teve bom resultado (0,67). O F1 final ficou em 0,29. Os resultados, possivelmente, foram prejudicados pela pequena quantidade de exemplos disponíveis.
\begin{table}[h]
\centering
\caption{Resultados de F1-score para o REN na CDJUR-BR e LENER-BR (C2, C3, C4 e C5) utilizando o modelo BERT}
\label{resultados c2-c5}
\begin{center}
\begin{tabular}{| c || >{\centering\arraybackslash} m{2cm} | >{\centering\arraybackslash} m{2cm} | >{\centering\arraybackslash} m{2cm} | >{\centering\arraybackslash} m{2cm} |}
\hline
\textbf{Entidade} & \multicolumn{4}{ c |} {\textbf{Cenário de Experimento}} \\
\cline{2-5}
& {\textbf{C2}} & {\textbf{C3}}& {\textbf{C4}} & {\textbf{C5}}\\ 
\hline
\hline
\textbf{JURISPRUDÊNCIA} &0.89&0.96&0.79&0.48\\
\hline
\textbf{LEGISLAÇÃO} &0.92&0.97&0.92&0.86\\
\hline
\textbf{LOCAL} &0.77&0.77&0.32&0.15\\
\hline
\textbf{PESSOA} &0.83&0.97&0.69&0.76\\
\hline
\hline
\textbf{F1-micro avg} &0.85&0.96&0.81&0.60\\
\hline
\textbf{F1-macro avg} &0.85&\textbf{0.92}&0.68&0.56\\
\hline
\textbf{F1-weighted avg} &0.85&0.96&0.79&0.74\\
\hline
\end{tabular}
\end{center}
\end{table}
\subsubsection{Análise dos resultados para os cenários comparativos com LENER-BR (C2, C3, C4 e C5)}
Os experimentos destes cenários foram realizados com o modelo baseado no BERT. Em uma comparação direta de C2 com C3, obsevamos que os resultados de C2 são inferiores aos obtidos em C3. Dito com outras palavras, os resultados do REN treinado e testado com a CDJUR-BR são um pouco inferiores aos resultados do REN quando se treina e se testa com o LENER-BR. A grande diversidade de documentos que compõem os \textit{corpus} da CDJUR-BR pode contribuir para o seu desempenho inferior (vide Tabela \ref{tabela-composicao-corpus}). Todavia, quando comparamos C4 com C5 (modelo treinado com a CDJUR-BR e testado com o LENER-BR comparado com modelo treinado com LENER-BR e testado com a CDJUR-BR), verifica-se que a CDJUR-BR obtém um desempenho muito superior (Média Macro F1 de 0,68 contra 0,56). Esse resultado pode indicar que a CDJUR-BR tem maior capacidade de adaptabilidade para reconhecer entidades de outro \textit{corpus} de documentos legais. A Tabela \ref{resultados c2-c5} apresenta os resultados obtidos para os cenários de 2 a 5 (C2 a C5).
\section{Conclusão}
\label{sec:conclusão}
Neste artigo, foi apresentada uma metodologia própria de anotação manual de documentos jurídicos, que serviu para criar a coleção de entidades nomeadas padrão ouro, chamada CDJUR-BR. A coleção é formada pelas classes semânticas Pessoa, Prova, Pena, Endereço, Sentença e Norma, dispondo de 44.526 anotações realizadas para 21 entidades nomeadas distintas. O processo de anotação foi cuidadosamente avaliado para garantir a exatidão e confiabilidade dos dados. Verificou-se que em 73\% dos documentos, a concordância entre anotadores alcançou coeficiente Kappa superior a 0,50. Ainda assim, os demais documentos passaram por revisões com especialistas e etapas extras de refinamentos de algumas classes semâticas. 

Realizamos experimentos na tarefa de Reconhecimento de Entidades Nomeadas com os modelos SPACY, BI-LSTM + CRF e BERT. Os resultados apontaram superioridade do modelo BERT com a Média Macro da Medida-F geral de 0,58, demonstrando a viabilidade da CDJUR-BR para o treinamento de modelos de aprendizado automático em aplicações de \textit{LEGAL AI}. Além disso, fizemos testes comparativos entre CDJUR-BR e LENER-BR. Embora a CDJUR-BR tenha atingido uma precisão menor do que o LENER no Reconhecimento de Entidades Nomeadas, pudemos verificar que a CDJUR-BR foi muito superior quando a tarefa foi reconhecer as entidades do \textit{corpus} LENER-BR.

\subsection{Limitações e Trabalhos Futuros}
\label{sec:limitacoes-trabalhos-futuros}

A partir dos experimentos realizados, percebemos que as características dos dados impactaram consideravelmente no desempenho dos modelos. Isso sugere melhorias para aumentar a quantidade de exemplos de algumas entidades para que se possa balancear algumas categorias de entidades e reduzir o impacto dos dados nos modelos REN. Os resultados também mostram que não existe um modelo REN universal que reconheça todas as entidades da melhor maneira. Portanto, um classificador formado por vários modelos poderia ser construído para alcançar melhores resultados em categorias ou entidades nomeadas específicas. Foi identificada mais uma limitação neste estudo ao comparar a capacidade da CDJUR-BR em reconhecer entidades do domínio jurídico em mais \textit{corpus}.

Em trabalhos futuros, planejamos melhorar a desambiguidade entre entidades como meio de aumentar o desempenho dos modelos. Realizar mais anotações objetivando aumentar as EN minoritárias e reduzir o desbalanceamento entre entidades.Desenvolver novos modelo REN a fim de selecionar aqueles de melhor desempenho para entidades específicas. E, realizar comparações com vários \textit{corpus} no domínio jurídico.

\bibliographystyle{unsrt}  
\bibliography{references}

\begin{thebibliography}{10}

\bibitem{peixoto2020projeto}
Fabiano~Hartmann Peixoto.
\newblock Projeto victor: relato do desenvolvimento da intelig{\^e}ncia
  artificial na repercuss{\~a}o geral do supremo tribunal federal.
\newblock {\em Revista Brasileira de Intelig{\^e}ncia Artificial e
  Direito-RBIAD}, 1(1):1--22, 2020.

\bibitem{kanapala2019text}
Ambedkar Kanapala, Sukomal Pal, and Rajendra Pamula.
\newblock Text summarization from legal documents: a survey.
\newblock {\em Artificial Intelligence Review}, 51(3):371--402, 2019.

\bibitem{yamada2019building}
Hiroaki Yamada, Simone Teufel, and Takenobu Tokunaga.
\newblock Building a corpus of legal argumentation in japanese judgement
  documents: towards structure-based summarisation.
\newblock {\em Artificial Intelligence and Law}, 27(2):141--170, 2019.

\bibitem{zhong2020does}
Haoxi Zhong, Chaojun Xiao, Cunchao Tu, Tianyang Zhang, Zhiyuan Liu, and Maosong
  Sun.
\newblock How does nlp benefit legal system: A summary of legal artificial
  intelligence.
\newblock {\em arXiv preprint arXiv:2004.12158}, 2020.

\bibitem{angelidis2018named}
Iosif Angelidis, Ilias Chalkidis, and Manolis Koubarakis.
\newblock Named entity recognition, linking and generation for greek
  legislation.
\newblock In {\em JURIX}, pages 1--10, 2018.

\bibitem{boella2019semi}
Guido Boella, Luigi Di~Caro, and Valentina Leone.
\newblock Semi-automatic knowledge population in a legal document management
  system.
\newblock {\em Artificial intelligence and Law}, 27(2):227--251, 2019.

\bibitem{yadav2019survey}
Vikas Yadav and Steven Bethard.
\newblock A survey on recent advances in named entity recognition from deep
  learning models.
\newblock {\em arXiv preprint arXiv:1910.11470}, 2019.

\bibitem{schmitt2019replicable}
Xavier Schmitt, Sylvain Kubler, J{\'e}r{\'e}my Robert, Mike Papadakis, and Yves
  LeTraon.
\newblock A replicable comparison study of ner software: Stanfordnlp, nltk,
  opennlp, spacy, gate.
\newblock In {\em 2019 Sixth International Conference on Social Networks
  Analysis, Management and Security (SNAMS)}, pages 338--343. IEEE, 2019.

\bibitem{li2020survey}
Jing Li, Aixin Sun, Jianglei Han, and Chenliang Li.
\newblock A survey on deep learning for named entity recognition.
\newblock {\em IEEE Transactions on Knowledge and Data Engineering},
  34(1):50--70, 2020.

\bibitem{jiang2016evaluating}
Ridong Jiang, Rafael~E Banchs, and Haizhou Li.
\newblock Evaluating and combining name entity recognition systems.
\newblock In {\em Proceedings of the Sixth Named Entity Workshop}, pages
  21--27, 2016.

\bibitem{atdaug2013comparison}
Samet Atda{\u{g}} and Vincent Labatut.
\newblock A comparison of named entity recognition tools applied to
  biographical texts.
\newblock In {\em 2nd International conference on systems and computer
  science}, pages 228--233. IEEE, 2013.

\bibitem{coan2003atributos}
Emerson~Ike Coan.
\newblock Atributos da linguagem jur{\'\i}dica.
\newblock 2003.

\bibitem{de2018lener}
Pedro Henrique~Luz de~Araujo, Te{\'o}filo~E de~Campos, Renato~RR de~Oliveira,
  Matheus Stauffer, Samuel Couto, and Paulo Bermejo.
\newblock Lener-br: a dataset for named entity recognition in brazilian legal
  text.
\newblock In {\em International Conference on Computational Processing of the
  Portuguese Language}, pages 313--323. Springer, 2018.

\bibitem{leitner2020dataset}
Elena Leitner, Georg Rehm, and Juli{\'a}n Moreno-Schneider.
\newblock A dataset of german legal documents for named entity recognition.
\newblock {\em arXiv preprint arXiv:2003.13016}, 2020.

\bibitem{huang2020named}
Wenming Huang, Dengrui Hu, Zhenrong Deng, and Jianyun Nie.
\newblock Named entity recognition for chinese judgment documents based on
  bilstm and crf.
\newblock {\em EURASIP Journal on Image and Video Processing}, 2020(1):1--14,
  2020.

\bibitem{oliveiraulyssesner}
Adriano~LI Oliveira.
\newblock Ulyssesner-br: a corpus of brazilian legislative documents for named
  entity recognition.

\bibitem{santos2006golden}
Diana Santos and Nuno Cardoso.
\newblock A golden resource for named entity recognition in portuguese.
\newblock In {\em International Workshop on Computational Processing of the
  Portuguese Language}, pages 69--79. Springer, 2006.

\bibitem{klie2018inception}
Jan-Christoph Klie, Michael Bugert, Beto Boullosa, Richard~Eckart de~Castilho,
  and Iryna Gurevych.
\newblock The inception platform: Machine-assisted and knowledge-oriented
  interactive annotation.
\newblock In {\em Proceedings of the 27th International Conference on
  Computational Linguistics: System Demonstrations}, pages 5--9, 2018.

\bibitem{cejuela2014tagtog}
Juan~Miguel Cejuela, Peter McQuilton, Laura Ponting, Steven~J Marygold, Raymund
  Stefancsik, Gillian~H Millburn, Burkhard Rost, FlyBase Consortium, et~al.
\newblock tagtog: interactive and text-mining-assisted annotation of gene
  mentions in plos full-text articles.
\newblock {\em Database}, 2014, 2014.

\bibitem{manning1999foundations}
Christopher Manning and Hinrich Schutze.
\newblock {\em Foundations of statistical natural language processing}.
\newblock MIT press, 1999.

\bibitem{silva2013transparencia}
Rosane Leal~da Silva, Patr{\'\i}cia~Adriani Hoch, and Lucas~Martins Righi.
\newblock Transpar{\^e}ncia p{\'u}blica e a atua{\c{c}}{\~a}o normativa do cnj.
\newblock {\em Revista direito GV}, 9:489--514, 2013.

\bibitem{mikheev1999named}
Andrei Mikheev, Marc Moens, and Claire Grover.
\newblock Named entity recognition without gazetteers.
\newblock In {\em Ninth Conference of the European Chapter of the Association
  for Computational Linguistics}, pages 1--8, 1999.

\bibitem{hovy2010towards}
Eduard Hovy and Julia Lavid.
\newblock Towards a ‘science’of corpus annotation: a new methodological
  challenge for corpus linguistics.
\newblock {\em International journal of translation}, 22(1):13--36, 2010.

\bibitem{mchugh2012interrater}
Mary~L McHugh.
\newblock Interrater reliability: the kappa statistic.
\newblock {\em Biochemia medica}, 22(3):276--282, 2012.

\bibitem{graves2005framewise}
Alex Graves and J{\"u}rgen Schmidhuber.
\newblock Framewise phoneme classification with bidirectional lstm and other
  neural network architectures.
\newblock {\em Neural networks}, 18(5-6):602--610, 2005.

\bibitem{hochreiter1997long}
Sepp Hochreiter and J{\"u}rgen Schmidhuber.
\newblock Long short-term memory.
\newblock {\em Neural computation}, 9(8):1735--1780, 1997.

\bibitem{lafferty2001conditional}
John Lafferty, Andrew McCallum, and Fernando~CN Pereira.
\newblock Conditional random fields: Probabilistic models for segmenting and
  labeling sequence data.
\newblock 2001.

\bibitem{devlin2018bert}
Jacob Devlin, Ming-Wei Chang, Kenton Lee, and Kristina Toutanova.
\newblock Bert: Pre-training of deep bidirectional transformers for language
  understanding.
\newblock {\em arXiv preprint arXiv:1810.04805}, 2018.

\bibitem{souza2020bertimbau}
F{\'a}bio Capuano~de Souza et~al.
\newblock Bertimbau: pretrained bert models for brazilian portuguese=
  bertimbau: modelos bert pr{\'e}-treinados para portugu{\^e}s brasileiro.
\newblock 2020.

\bibitem{spacy2}
Matthew Honnibal and Ines Montani.
\newblock {spaCy 2}: Natural language understanding with {B}loom embeddings,
  convolutional neural networks and incremental parsing.
\newblock To appear, 2017.

\bibitem{ramshaw1999text}
Lance~A Ramshaw and Mitchell~P Marcus.
\newblock Text chunking using transformation-based learning.
\newblock In {\em Natural language processing using very large corpora}, pages
  157--176. Springer, 1999.

\end{thebibliography}

\end{document}